# Unsupervised Foveal Vision Neural Networks with Top-Down Attention

Ryan Burt, Nina N. Thigpen, Andreas Keil, Jose C. Principe

Computational NeuroEngineering Laboratory
Department of Electrical and Computer Engineering University of Florida
Gainesville, Florida 32601

*Abstract*—Deep learning architectures are an extremely powerful tool for recognizing and classifying images. However, they require supervised learning and normally work on vectors the size of image pixels and produce the best results when trained on millions of object images. To help mitigate these issues, we propose the fusion of bottom-up saliency and top-down attention employing only unsupervised learning techniques, which helps the object recognition module to focus on relevant data and learn important features that can later be fine-tuned for a specific task. In addition, by utilizing only relevant portions of the data, the training speed can be greatly improved. We test the performance of the proposed Gamma saliency technique on the Toronto and CAT2000 databases, and the foveated vision in the Street View House Numbers (SVHN) database. The results in foveated vision show that Gamma saliency is comparable to the best and computationally faster. The results in SVHN show that our unsupervised cognitive architecture is comparable to fully supervised methods and that the Gamma saliency also improves CNN performance if desired. We also develop a top-down attention mechanism based on the Gamma saliency applied to the top layer of CNNs to improve scene understanding in multi-object images or images with strong background clutter. When we compare the results with human observers in an image dataset of animals occluded in natural scenes, we show that top-down attention is capable of disambiguating object from background and improves system performance beyond the level of human observers.

## I. Introduction

Neural networks and deep learning (DL) architectures are the current state-of-the-art for image classification and recognition. They have been shown to reliably distinguish between as many as 10,000 different classes of objects [1]. These networks, however, currently fall well short of human capabilities in two areas: recognizing objects based on a relatively small number of examples and localizing and detecting multiple objects in a single scene. One of the culprits is that computer architectures rasterize the image into a long vector with a size given by the number of pixels in the image. This approach, universally accepted as the standard, simplifies the processing in neural network algorithms but it is a brute force procedure that even loses local information.

In order to move towards more autonomous vision systems, we need an architecture that can extract features from a wide range of objects in cluttered scenes with no labels in training. Humans have the remarkable ability to view a scene and form an overall representation in a very short time [14]. However, due to the complexity of visual scene understanding, it is reasonable to assume that humans do not process an entire scene at once, or even fixate on and process every small region in an image. Instead, the human vision system (HVS) consists of a number of brain areas that operate in a massively parallel fashion, and which often are grouped in two subsystems based on neuroanatomy [2][11] [18]: one for object recognition and one for spatial localization. The ventral stream, or "what" pathway, consists of visual areas V1, V2, V4, and continues to the inferior temporal cortex. Among other functions, it performs neurocomputations for identifying and semantically representing visual objects [13] [17]. The dorsal stream (Figure 1), or "where" pathway, goes through V1, V2, the dorso-medial area to the posterior parietal cortex where object locations in internal coordinates are computed, for example to enable control of eye movements (saccadic changes of gaze location) by the oculomotor system [2]. Humans saccade to a new fixation location at an average 3 times a second, varying with task demands and complexity of the visual scene [9]. This is adaptive because the combination of optical and neurophysiological features in the visual pathway yields a full resolution of the scene only for the foveal and parafoveal portion of the visual field, corresponding to approximately 4 to 5 degrees of the visual angle [7], [17],[21]. Thus, saccades enable the brain to sequentially sample information from the full image field, using high acuity foveal vision [22]. Recently, it has been proposed that the initiation of a saccade is also the beginning of a visual processing cycle aimed at actively sensing the environment, and ultimately recognizing objects in the ventral pathway [15], very much like sniffing for odors. The advantage of this complex sensory motor coordination is to reduce the substantial complexity of the global visual scene to a series of simpler perceptual decisions made at the local level. Saccade control during active exploration involves the entire brain, in a dynamic complex process that is not fully characterized [3][6]. However, there is agreement that two processes are at play: a bottom-up process that selects "interesting" local patches based on their saliency salient, i.e. sharp change of luminance, contrast, color or texture in a local region of the image [7]; and a top-down process that guides the eye to relevant visual details to disambiguate the scene with respect to the current goal [10]. These two processes are commonly called overt visual attention [12] and eliminate the need to operate with the full visual scene to save limited resources, and to facilitate the computation of motor output.

Here, we propose to design a fovea-based image processing system that is inspired by the HVS incorporating two pathways and an attention module including both bottom-up and top-down saliency. The foveal image patch will be delivered to a redundancy reduction object recognition algorithm [31] which extracts features in a self-organizing way (i.e. without labels) and stores them in an external memory for future use. The key characteristics of our approach relate to the algorithmic pipeline, which is controlled by foveation; the construction of a video stream with fixed retina to improve the unsupervised recognition of objects in the foveated patch; and the inclusion of top-down attention to modify the order of the foveated patches according to the analysis goal.

The paper is organized as follows: Section I reviews the literature, section II presents the overall architecture and discusses primarily the bottom up and top down focus of attention, while section III presents results with the two attention mechanisms in both synthetic data, larger data sets, and a comparison with human subjects observing highly cluttered images. The paper end with a conclusion.

## II. Related Work

### A. Attention Systems

One of the hallmarks of foveal vision as discussed above is that is highly dependent on the coordination of separate streams for object recognition and selection of salient areas in the image space. Not even the brain has enough bandwidth to understand the continuous 180 degrees video stream that reaches the eyes. Hence, it sparsifies selectively and intentionally the video input for real time interaction with the world. By using this divide-and-conquer approach, the brain can quickly process large, unwieldy images and break them into smaller pieces that require the more intense computation required to extract and recognize object features from an image. This patch selection approach is compatible with the fact that the eye can only perceive the world in high resolution in a narrow cone (5 degrees) centered at the fixation point. The selection of where to look (the saccades) involves the entire brain, e.g. executive, associative and motor areas (Figure 1) and is rather complex. The discovery of simplifying assumptions for bottom-up and top-down processes are enabling inspirations for machine learning architectures, as discussed below.

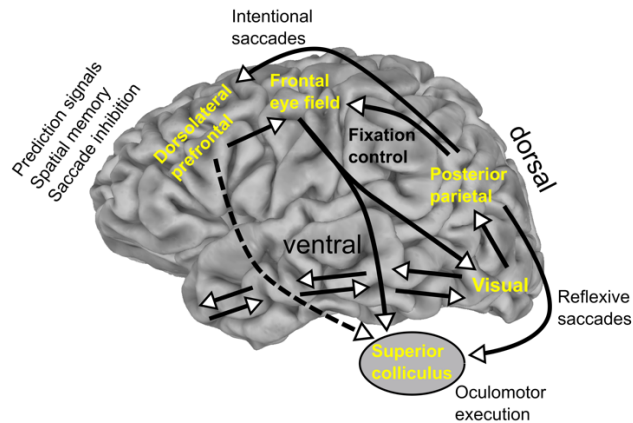

Figure 1. The "what" and "where" paths in the human visual system

Bottom-up Attention: The HVS dual-stream organization introduces new problems: finding the regions that contain relevant information, i.e. implementing visual attention, on top of recognizing the patch content. Recently, machine learning methods have been proposed to suggest regions both within the structure of the network [24] or as a separate mechanism based on image features [25]. These approaches are unfortunately supervised and are trained to choose regions that contain data most relevant to the label. Since the introduction of Itti's method in 1998 [26], saliency has become a popular way to predict visual attention in images, which could therefore be used to segment out the interesting image regions for faster processing. Saliency is defined as the state or quality by which an object stands out relative to its neighbors. An object tends to be more salient if it is brightly colored, flashy, and altogether different from its surroundings. Employing saliency as a proxy for visual attention enables the design of unsupervised systems that can: (1) quickly select regions of interest; (2) dedicate resources for more computationally intensive processing only to the selected regions; (3) combine these representations into an overall understanding of a complex scene in much the same way the HVS works. Practically, the human visual cortex must remember and infer parafoveal and peripheral information, or use a combination of the two, to estimate regions of interest for future fixation locations. As shown from empirical research on saccadic exploratory eye movements, these future fixations will target the regions in the visual periphery [3] [5].
Most saliency measures work by combining a number of simple features such as color, intensity, and orientation to find distinct regions in images that could attract the human eye. Three competing views of saliency are the center-surround methods that compare a local center to a neighborhood [26], [34], [35], [36] [37], the global context methods that compare regions to other regions from any location in the image [38], [39], and the normal image methods that compare an image to a standard ideal [40], [41], [42], [43], [44], [45]. Saliency metrics have been used in an effort to reduce computation in image and video processing, often in lossy compression algorithms that keep high resolution data only in salient areas [46] [47].

<u>Top-down saliency:</u> It turns out that bottom-up attention does not explain all the visual attention mechanisms discovered in human perception. Cognitive scientists have empirically demonstrated that when the same individual views the same scene twice, she/he will change their saccadic exploration paths [4] [21]. Therefore, this indicates that there is also a top-down attention process from the executive cortex (frontal) that changes visual processing, by conditioning the extraction of relevant information from the scene. Conceptually this selection is also based on some sort of saliency, but instead of working on image pixels, saliency is applied to more abstract set of knowledge representations. The details are far from being fully understood [19] [22], but the existence of deep learning architectures and its multiscale learned representations opens the door for experimenting with saliency algorithms on their top layers. It has been experimentally verified [25] that deep learning networks tend to cluster features of specific objects in their top layers. Hence, we can attempt to apply similar saliency algorithms, not to the pixels but to the feature maps created by convolution neural networks (CNNs). The Top-Down attention becomes another input to the system that can change the priority of the search in image space, by modifying the bottom-up saliency maps.

### B. Object Recognition

With saliency functioning as guided search, the second pathway will replicate the ventral stream and will form representations of and extract features from objects. The ventral stream receives visual data from the fovea and builds a representation through the visual cortex that is then sent to a temporary (working) memory to enable inference about the scene composition. Finally, objects that are relevant for the state of the subject are permanently stored in a visual long-term memory. It is well accepted that human memory is content addressable [64], which is very efficient because it utilizes the metric of the internal representations, instead of an address bus as in our digital computers. Normally in computer vision architectures, there is no use of external memory blocks. A neural network builds its own internal long-term memory of the input data in its parameters through learning, or a short-term memory in its recurrent connections, like LSTMs [23]. But this internal memory is not shared with other modules in the architecture. Therefore, we propose to implement explicitly content addressable memories in our architecture. With regards to object recognition, we will be using a deep learning neural network. But notice that by focusing the representation on only foveated patches, rather than the entire scene, we save computation and may even improve recognition because background and other objects are distractors. By segmenting objects around highly salient points found in the attention mechanism, we can restrict the role of the network to find invariant representations of the objects in isolation.

These deep networks are generally trained on large datasets such as Imagenet [1] or MNIST [27] that contain tens of thousands up to millions of labeled images. By backpropagating the errors in the class labels through the network, the network is able to learn the relevant features for predicting the label associated with the image. However, this learning becomes harder when multiple objects are contained within each image, each with its own label. In addition, supervised training requires labels for each image, which requires curating these large datasets and hampers their ability to be implemented outside of certain situations.

Training deep learning architectures without explicit class labels has been a growing area of research [48] [29]. In an effort to expand these techniques beyond datasets that come with an excess of labeled examples, there have been effort toward learning features based on other forms of supervision such as temporal and egomotion. Goroshin et. al [28] and Wang and Gupta [29] learned short term dependencies between subse- quent frames in video. Agrawal et al. modeled the egomotion of the camera in order to provide a form of supervision other than labels [30].

A different approach uses the input data as the desired response, i.e. it creates a generative model of the system that created the data. Helmholtz in the nineteen century wrote that the role of the visual system was exactly to model the external sources that created the sensory stimuli [8]. These techniques normally employ redundancy reduction techniques to extract the generative model that created the data. The deep predict coding network (DPCN) by Chalasani and Principe [31] [51] uses temporal predictions to learn in an unsupervised manner features through time and build object representations of video streams. This work was later extended to the recurrent winner take all autoencoder (RTWA) [52], which use a dual-stream autoencoder structure with a recurrent bottleneck layer to represent the current frame and predict the next frame (Figure A in the appendix). Both methods are self-organizing generative models, but use different algorithms: the DPCN uses a naïve Bayes approach to maximize free energy on a distributed multilayer topology, while the latter uses more traditional machine learning methods (stacked autoencoder and recurrent neural networks). This framework can be easily integrated to learn features from unlabeled image datasets by taking advantage of the sequence structure introduced through a prescribed movement of a retina (creating a video) covering the saliency patch [65].

Despite these recent advances in image processing, classification results on image datasets with multiple objects in complex scenes has stagnated when using only convolutional methods. Recently, research has begun into breaking images down into regions, then performing classification on these rather than the entire image [32]. By fusing these region detection algorithms with the recent advances in convolutional networks, classification performance on datasets such as VOC2012 have improved by up to 30% [25].

Most of the region classification methods proposed at this time were designed to be trained in conjunction with deep convolution networks, such as OverFeat [33]. OverFeat consists of a single convolutional network that is applied at multiple locations via a sliding window before producing a distribution that predicts the bounding box containing the targeted object. Alternatively, the R-CNN uses a separate region proposal method (selective search), before separately sending these regions to a CNN for classification and then finally recombining similar regions [25]. Despite the different paradigms, processing smaller regions of images has the potential to be the next breakthrough in computer vision by reducing the brute force sliding windows in the CNNs. The Spatial Transformer Network (STN), on the other hand, integrates a differentiable image transform into the overall network structure that is capable of learning which features in an image best discriminate objects by their labels, focusing in on these objects accordingly [24]. More recently visual question answering (VQA), the bottom-up mechanism (based on Faster R-CNN) proposes image regions, while the top-down mechanism determines feature weightings [66]. The stacked what-and-where autoencoder is a different approach of implementing divide and conquer in a deep architecture [67] without supervision, and exploits the same discriminative-generative principles of our approach. Other applications of foveation in DL are the recent DeepFovea by Facebook to speed up and bring low-power to rendering in virtual reality environments [68], and to implement gaze prediction with DL [69], DeepFix [70] or efficient egocentric machine perception [71], or for aesthetic image appreciation [72].

## III. METHODS

Humans experience the world's static scenes through movement, whether by moving fixations across a painting or walking around a still landscape. This motion is inherent to understand our environment; despite the lack of change in the physical properties of the scene, the information sent to the visual cortex through the eyes is constantly changing at a slow pace as the viewpoint is updated. The temporal coherence builds the full understanding of the scene as objects are recognized and placed into working memory as the brain searches out new fixations. Time disambiguates space.

### C. Proposed Architecture

As in HVS, we propose to merge information from separate paths in a machine learning architecture: one path for attention, which selects visually informative regions (bottom-up and top-down saliency modules), and one for representing, organizing and storing extracted objects in an external long-term memory module. The bottom-up saliency module will initiate and drive the full vision system operation in cycles, started by foveation to the center of the image, to select saliency patches and forward them to the object recognition module one at a time. Since, one image may have more than one salient patch, the process of selecting salient regions needs to be repeated until the saliency map is featureless, achieved by an hyperparameter to define what is "sufficiently different" from background. Meanwhile, the extracted objects must be temporary stored in an internal canvas, a.k.a working memory, that summarizes the visual scene, and can be used for further inference. This partitioning of image information to simplify model building, brings with it many modifications in the standard DL architecture that has not been fully exploited in previous works [1], [61]. The hallmark of our architecture is the use of unsupervised learning in an acquire-process cycle, combination of bottom up saliency with top down attention and storage of extracted objects in external memories. Figure 2 shows our proposed architecture.

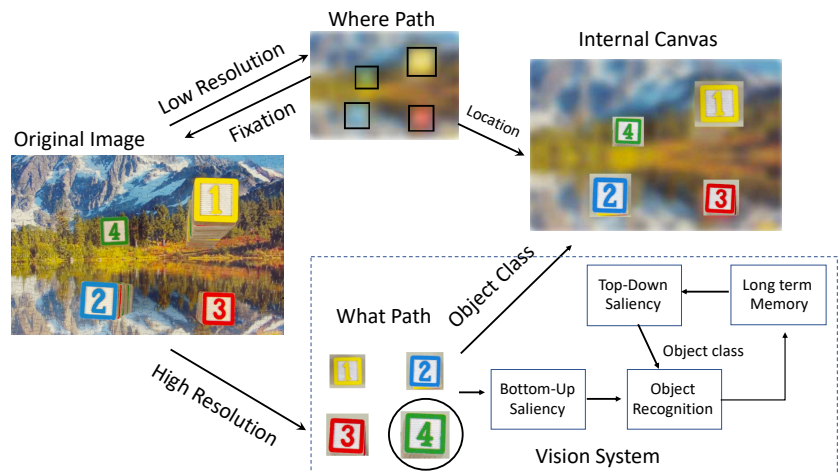

Figure 2. Within the framework of the what-where paths, we present our vision system architecture using sequential patch processing for scene understanding. The "where" path is implicitly included by the search for salient image patches (bottom-up saliency), as well as by the top-down attention mechanism. Note that the external memory to the object recognition module is also needed for inference and storage of object identity.

Each of the salient image patches are sent, one at a time, to the recognition module where the patch content is analyzed with a self-organizing algorithm to represent or recognize a possible object in each patch. The top layer of the recognition system (DPCN) or the codes of the bottleneck layer in RTWA are then stored in an external content addressable memory (CAM) for future use. After training and during testing, the system still has the ability to speed up recognition of a scene if it is instructed by the user (or the task itself) to search a certain object type in the scene. Currently, the top-down attention module can modify, as a prior, the bottom-up module with the characteristics of the preferred object. We will explain each one of these modules below.

### A. Gamma Saliency

An effective attention mechanism should meet a few basic requirements to bring two pathways into a single vision system pipeline. First, the calculation should be done quickly so that the attention works to speed scene recognition, not slow it by compounding the data. Second, the bottom-up attention system should trigger the recognition module, not in reverse order, i.e. be driven by recognizing objects and then assigning saliency scores. Since the dorsal stream of the HVS uses the peripheral, and therefore blurred, vision as the input to determine fixations, the system should be able to work only with low-level features.

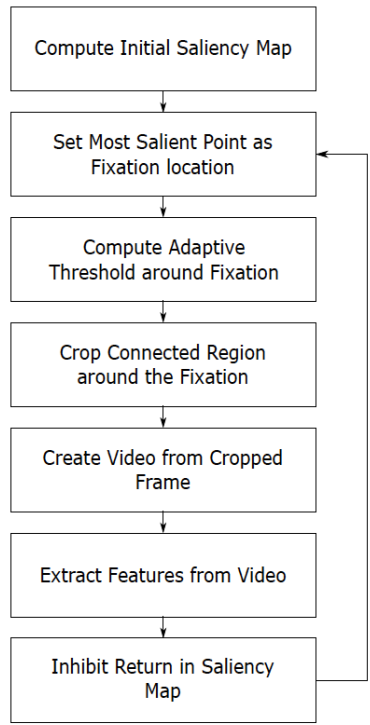

Figure 3. The cycle of acquiring and processing information in our architecture, initiated with the saccade to the center of the image.

To accomplish this list, we use a simple center surround saliency method that computes local differences in regions at different scales. Although high level saliency methods exist which predict human fixations very well, these often require extensive training, require full object recognition, and are slower to compute than the more classic signal processing methods. We adopt here the Gamma kernels, which have been used for target detection [53]. The circular shape of the kernel is ideal for comparing a center region to a local neighborhood, and the size of the saliency patch can easily be controlled to fit the object size through two parameters (the decay $\mu$ and the order *k* of the Gamma function), which allows for easy change in spatial scales.

Similar to the Itti method and others, the Gamma saliency [37] is based on the center surround principle: a region is salient if it is sufficient different from the surrounding neighborhood. In order to compute these local differences, we use a 2D gamma kernel that emphasizes a central area while contrasting it with a local neighborhood through convolution:

$$g_{k,\mu}(n_1, n_2) = \frac{\mu^{k+1}}{2\pi k!}\left(\sqrt{n_1^2 + n_2^2}\right)^{k-1} e^{-\mu\sqrt{n_1^2+n_2^2}} \qquad (1)$$

For this kernel, $n_1$ and $n_2$ are the local support grid, $\mu$ is the shape parameter, and $k$ is the kernel order. Using $\mu$ and $k$, we can control the shape of the kernel: when $k = 1$, the kernel peak is centered around the origin of the patch (exponential decay). For kernel orders $k > 1$, the peak is centered at $k/\mu$ away from the origin. In addition, smaller values of $\mu$ will increase the width of the peak. With these two parameters we can construct a 2D circularly symmetric shape that compares a center region to a surrounding neighborhood by subtracting the kernel with order $k > 1$ from the $k = 1$ kernel. The peak of the 1st order kernel functions as the center of the area to be tested, while the higher order kernel forms the surrounding neighborhood. By adjusting the shape parameter and order of the ring kernel we can control the size and location of the neighborhood relative to the center, as well as adjust the size and location of the neighborhood relative to the center. The 2D kernel is them moved over the full image, with a stride given by twice the radius of the ring kernel.

One of the advantages of the Gamma kernel is that it easily allows for the estimation of the object size, by utilizing multiple rings and implementing successive subtractions over consecutive rings. For a multiscale saliency measure, we simply combine multiple kernels of different sizes before the convolution stage (2). The scale parameter affects all of the kernels in the same way, i.e. a kernel with a larger center scale is subtracted by a surround kernel with a larger and further removed neighborhood, effectively searching for larger objects. By summing all the kernels before the convolution stage, we create a system which is capable of computing saliency at different scales adding little extra computation beyond a simple summation. The kernel summation is described in (2), where all $k$ for even $m$ are 1 to create the center kernels. The number of different scales is half the value of $m$.

$$g_{total} = \sum_{m=0}^{M-1}(-1^m)g_m(k_m, \mu_m) \qquad (2)$$

In this work we do not take advantage of the recursive computation of the gamma function, because we precompute the kernel. However, if one is interested in space time saliency, the recursive computation of the gamma function is very convenient.

Apart from the local functions, the rest of the saliency measure is constructed similarly to the other center surround methods [49]: the image is broken into local feature matrices of predetermined size, each matrix is convolved with the multiscale kernel, the matrices are combined and exponentiated to accentuate peaks, then post processing is performed to boost results using a Gaussian blur and a center bias.

The feature matrices to compute the saliency are composed in the CIELab color space, which has three matrices - one luminence matrix and two color opponency matrices. In CIELab space, the distance between two colors can be calculated using the Euclidean distance, which is a useful property that we take advantage of in the convolution. Each of these matrices is convolved with the multiscale gamma kernel to get the saliency measure in each channel (3). In the following equations, ∘ is the convolution operator.

$$S = \frac{|g \circ L| + |g \circ a| + |g \circ b|}{3} \quad (3)$$

Once we have the overall combined saliency map, there are a few common postprocessing mechanisms used to improve results [55]. First, the main peaks in the measure are accentuated by raising the combined map to a power $\alpha > 1$. Next, it is well known that humans tend to fixate on the image center, so a Gaussian weighting is applied to the center of the image where the variance of the Gaussian is dependent on the image size. Finally, to reduce the noise effects and create a more streamlined map, the map is blurred using a small Gaussian kernel (4) as

$$S = \left(S^\alpha G_{\sigma^2}(.)\right) \circ G_{0.5}(.). \quad (4)$$

*B. Foveated Imaging*

There are fundamental differences between how saliency measures are tested and how the human vision system uses saliency to direct attentive exploration of the surrounding scene. Since human vision only has access to full resolution in the fovea area, the saliency metrics to properly mimic the human vision system must be able to find regions of interest in low resolution, outside the initial focal area. Interestingly, studies have shown that initial full previews of the scene can often hinder relevant object detection, meaning that the blurred initial glimpse can be an improvement over knowledge of an entire scene a priori [73]. However, saliency algorithms applied to digital images have per definition access to the full resolution across the field of view. To address this crucial difference between biology and computational study, we use foveated imaging, which uses images with a clear field of focus and a blurred periphery to mimic the HVS. Foveated imaging has been used mainly for compression and faster processing [74], [47]. In addition, some saliency metrics have been tested in multi-resolution images in an attempt to speed computation and improve results [75], [46], but study in this area is still limited. Currently the Lytro Illum camera, a light field camera [76], can be used for this purpose. But for comparisons with data sets in the literature, we will need to create foveated imaging by software.

To mimic the effect of the fovea, we created images that are increasingly blurred around a small high-resolution area (artificial fovea), employing the fast method developed by Geisler and Perry for images and videos in 2002 [77]. This method creates a variable resolution map around a center point (either pre-selected or input in real time by the user). The map is composed of a multi resolution pyramid creating by first blurring the original image with a small kernel (such as 3x3), then down sampling and blurring with the same kernel, then repeating the process to create 6-7 layers. These layers are blended with weights corresponding to the distance from the center point, thus creating the newly foveated image. The foveation mechanism contains a resolution parameter that controls the distance weights, which in turn affect both the size of the fovea and the amount of blur in the periphery. Figure 4 shows an example of the gamma saliency on a foveated image in the Toronto data set, along with the human fixation map.

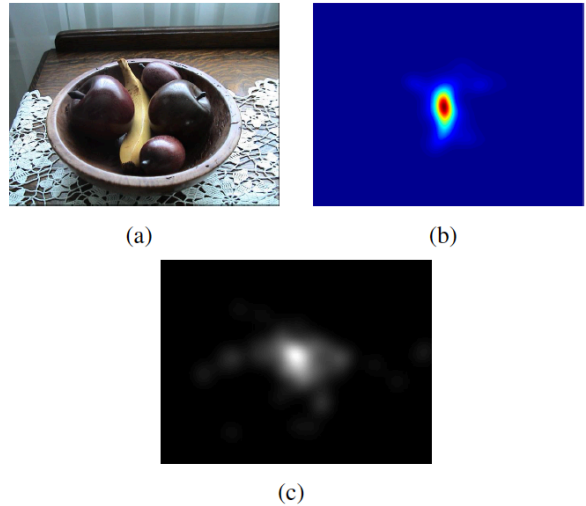

(a)                      (b)

(c)

Figure 4. Example image from the Toronto Saliency Dataset (A), saliency map produced by Gamma Saliency (B) and the ground truth fixation map (C).

## C. RTWA

Rather than using explicit labels in the form of class supervision, our vision module uses architectural constraints along with the structure inherent in a video stream in order to extract robust features from images. Here we use the RTWA [52], which uses a combination of a feedforward convolutional autoencoder and an RNN on the bottleneck layer that encodes a dynamic state that describes the change between consecutive frames. By using the same decoder at the end of each stream, the representations are forced to project to the same space and the error can be minimized. The cost function for the RTWA is given by

$$L_t = \mathbb{E}[(x_{t-1}D(E(x_{t-1}))^2 + (x_t D(R(x_{t-1}))^2] \quad (5)$$

where $x_t$ is the video stream at time t, the stateless encoder is $E$, the shared decoder is $D$, the RNN is $R$, and $E$ denotes the expectation operator. The architecture is trained using backpropagation through time and the architecture details are in the appendix.

## D. Combining Foveated vision with RTWA

The Gamma Saliency works on still images while the RTWA uses both spatial and temporal context to form images representations. In order to successfully take advantage of the temporal benefits of the RTWA, the attention mechanism must provide not only a salient point on which to focus, but a structured series of frames encompassing the object. There are multiple techniques that could be used for creating a sequence of frames from a patch of an image: first define the retina's size (the frame), and then scan the retina over the saliency patch, either in a circular or zig-zag path fashion. Both provide reasonable results; the important aspect is that the scan must be kept constant such that the RTWA's learned representations don't change across the frames from training to testing. This sequence of frames leads to a more invariant set of features learned by the vision system when compared to a simple cutout of pixels, as classified by a CNN [65].

The fine details for creating these videos is outlined in the fovea cycle of Figure 5. Given an input image (Fig 5A), compute the saliency map and the fixation is selected via the most salient point (Fig 5C). Since Gamma Saliency is a multi-scale measure, the underlying rings contain information to estimate the object extent, which leads to object segmentation matched to its size (Fig 5E-F). Working from this base, we fine tune the saliency patch to avoid nearby objects and recompute saliency and crop a tighter patch around the object (Fig 5G), create the video according to the scan selected (Fig 5H), and send it to the RTWA for representation and/or recognition. After this, the saliency map is locally inhibited by applying an inverted Gaussian that corresponds to the extent of the foveal area and with an amplitude estimated from the saliency value (Fig 5I). Once the video of the object has been created, the fixation is moved to the next most salient point, the next object viewed and sampled, and the process repeated.

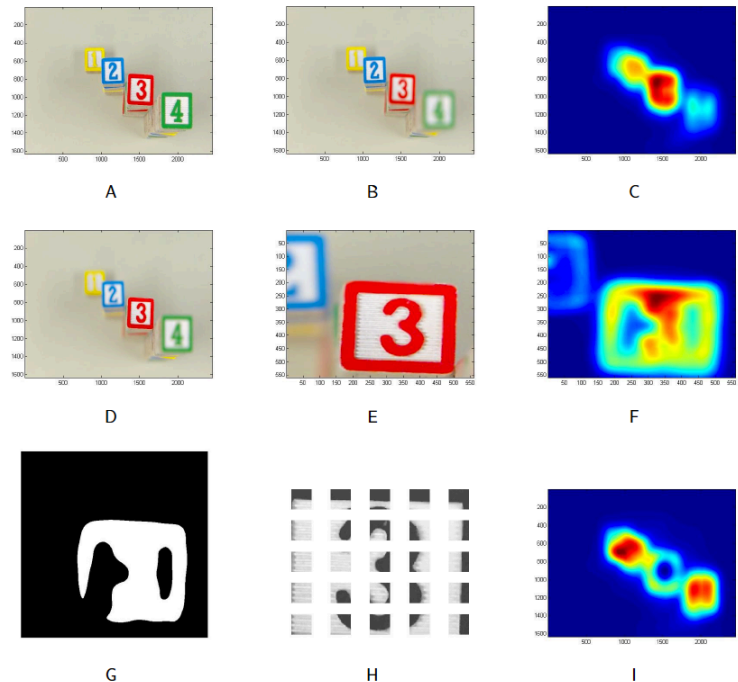

Figure 5. A series of images showing the progression of the focus of attention algorithm. A) The original image. B) The image focused on the center point. C) Saliency map created from center-focused image. D) The image refocused on the most salient point. E) The local patch containing the point. F) Local saliency map. G) The segmented object. H) A set of scanned frames. I) Saliency map around the new focus point with Gaussian inhibition at previously scanned locations.

*E. Top-Down Saliency Module.*

The last module of the architecture is the top-down visual attention, which implements a goal driven approach to guide the selection of scene objects. Top-down saliency is an extra input that can be used to modify the operation of a trained network to meet some other constraint. In the HVS, the executive cortex may want to direct the visual cortex to search for a piece of information needed to complete an inference. In machine learning, example applications are video question answering, as well as to find occluded objects in scenes. Current solutions in these domains [66] can still be largely improved. We propose a visual search module based on the same Gamma Saliency as detailed above. This is a center-surround method that takes advantage of the gamma kernel to compute a fast, multiscale contrast index. In bottom-up Gamma Saliency, the features maps were the channels of a LAB image. However, to implement top-down Gamma Saliency, we propose to use a set of feature maps $C_n$ from a fully convolutional network, where *n* is the number of feature maps [78]. Unlike fully-connected layers, convolutional layers of a neural network are agnostic to the size of the input. Therefore, it is possible to train a classification network on a standard dataset (such as MNIST), then strip the classification layers from the network and use their *N* internal outputs to preprocess images of any size, and for any purpose related to discrimination, which is the purpose of the training. In doing this, a set of feature maps $S_i$ are created that can be used to distinguish between *i* objects in the training set, i.e.

$$S_i = \frac{\sum_{n=1}^{N} w_n^i |g \circ C^n|^\alpha}{N} \quad (6)$$

where α is an enhancement parameter and ∘ means convolution. The set of weights $w_i$ can be learned on these maps to bias the saliency measures towards a specified target object. Since the feature maps were trained as part of a classification network, the features contained in these maps have already been optimized to discriminate objects in the image set. These saliency coefficients $S_i$ when utilized in conjunction with the bottom-up saliency templates will modify by multiplication their relative saliency value and lead to ranking of objects according to a goal. This will then decrease the number of fixations necessary to find a particular object, instead of using the order of the relative saliency in the scene that only uses pixel information. Of course, here we are using a simple weighted sum over the class of objects, so the discrimination achieved is not the best, but it is used here for simplicity and to test the approach.

We propose to learn the weights $w_n$ for each object *i* as follows: Need to use fovea vision to individualize which object is being observed. Over a training set with a specific object class selected, we can compute each raw saliency map $S_i$ with equal weights, i.e. assuming each $w_n^i = 1$. We then calculate the ratio of the raw saliency inside the bounding box $S_{I_n}^m$ to the raw saliency outside the box $S_{O_n}^m$ and find the mean value across the training data, where *m* is the specific image and *n* is the corresponding feature map, i.e.

$$w_n = \sum_{m=1}^{M} \frac{S_{I_n}^m / S_{O_n}^m}{M} \quad (7)$$

This value gives the average weight for each object across feature maps. After weighting each map in Eq. 7, we process and combine them in the same manner as bottom-up Gamma Saliency (Eq. 3). By processing them in this way, we end up with a new saliency that can bias the original bottom-up saliency towards certain objects, giving them a higher saliency than they would have in a pure bottom-up metric. We tested the method in the multi-object MNIST with very good results [65]. This method is similar to the work of Frintrop et. al [79], except that in our approach the feature maps are extracted from a network trained on the scene objects, instead of using pre-defined feature maps designed by hand. The procedure creates a weight matrix *W* that contains weights for each feature map and each class. Feature maps that always activate for a certain object are given a high weight, while feature maps with fewer activations are given a lower weight. Hence, each entry in this weight matrix corresponds to how highly each specific feature map activates for each class. Using these, we can quickly process an image and gain an idea as to the location of a given object.

IV. RESULTS

This section presents the experimental results. First, we will demonstrate the results of the gamma saliency when compared with other techniques. Then we apply foveated vision to the Street View House Numbers (SVHN) dataset to access the performance of the method in a realistic environment. Finally, we will show results and comparison with human observers on an occluded image dataset to show the importance of the top-down saliency. Information on the parameters used for the attention mechanism and the classifiers are presented in the appendix.

*Validation of Gamma saliency.*

Remember that our goal is to develop a methodology that uses only low-level features in the periphery (low resolution), and the goal is to be fast to compute, and agree with the human foveation. Therefore, we ultimately compare the methods with respect to human eye tracking. Saliency can be thought as object detection, and as such it is important to use detection theory as the underlying theory to compare different saliency detectors. To compare this new saliency metric with other common methods, results were computed on the Toronto dataset [80] and the CAT2000 training database [81]. The Toronto database consists of 120 images shown to 20 students for four seconds of free viewing. The CAT2000 database has 2000 images drawn from 20 different categories for a wide variety of image foregrounds and backgrounds, as well as the fixation data from 18 different observers. The observers were given the task of free viewing each image for five seconds with one degree of visual angle corresponding to roughly 38 pixels in each image. Each set of saliency maps were computed with the default set of parameters recommended by the algorithms.

For Gamma Saliency, the parameters used were $k = [1, 26, 1, 25, 1, 19]$, $\mu = [2, 2, 1, 1, .5, .5]$, and $\alpha = 5$. This gives center surround differences at three scales, as in [82], set to neighborhood sizes of 13, 25, and 38 pixels. All images are resized to 128x171 to speed processing time. $\alpha$ was selected by performing a grid search on the integers between 1 and 20. The maps were then compared to the collected fixation data using the following five metrics: the area under receiver operating characteristic (ROC) curve created by Judd [83], the area under ROC curve by Borji [84], the similarity measure [85], the correlation coefficient, and the normalized scanpath saliency [86]. The area under ROC curve by Judd is measured as the proportion of saliency map values above a threshold at the fixation locations to the number of values below the threshold at the fixation locations. In contrast, Borji's version of the area under ROC curve measure the proportion of true positives to false positives, which are the values in the saliency map above a threshold that do not correspond to a fixation location. The similarity measure treats each map as a distribution and computes the histogram intersection. The correlation measure is Pearson's linear coefficient between the two maps. Lastly, the normalized scan path saliency refers to the mean value of the normalized saliency map at fixation locations. In each of the metrics, the higher number indicates a better result. Also, note that these metrics only deal with finding the location of the fixation, not determining what the object is or its size.

To estimate the computation time, each algorithm was set to produce a saliency map sized 128x171 to ensure that algorithms that down sample don't have an inherent advantage for computation time. All times were computed on PC running Matlab R2012a on an i5-2310 clocked at 2.9GHz. Fig 6 shows the ROC curves for different scales. Table 1 shows the full results from comparing the saliency maps with the fixation maps in the Toronto database across five different metrics along with the mean time to create a saliency map from a single image in the database, with the best results for each metric in bold. Fig 7 shows the ROC curves calculated with the Judd method for each metric. Gamma saliency performs the best in four of five metrics, with the closest competitor being GBVS. Gamma saliency is also the fastest since it is based on a convolutional filter. Table 2 shows the results for the CAT2000 database. Once again Gamma saliency performs the best in 4 of 5 metrics and computes the saliency maps in the fastest times.

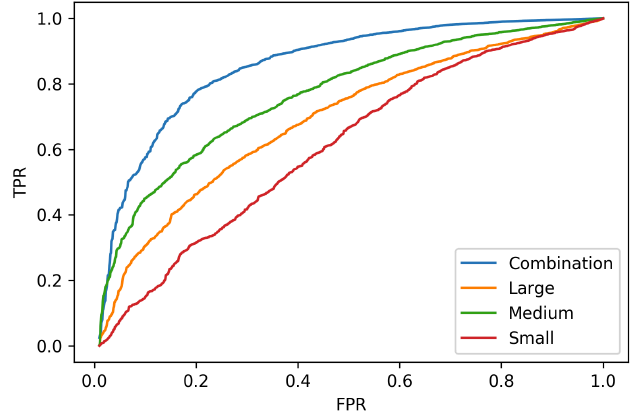

Figure 6. ROC curves for different scales of gamma saliency on the Toronto Saliency Dataset.

Table 1. Attention Prediction Results on the Toronto Database

| Method \ Metric | ROC (Judd) | ROC (Borji) | Similarity | Correlation | NSS | Time (s) |
| --- | --- | --- | --- | --- | --- | --- |
| Itti | .712 | .597 | .384 | .275 | .341 | .280 |
| AIM | .746 | .632 | .403 | .363 | .479 | 1.10 |
| Torralba | .684 | .600 | .374 | .292 | .360 | .78 |
| GBVS | .848 | .677 | .488 | .570 | .638 | 1.03 |
| FES | .847 | .586 | .520 | .572 | .446 | .21 |
| RARE2012 | .785 | .625 | .477 | .551 | .489 | 1.39 |
| RCS | .747 | .609 | .431 | .414 | .413 | 15.84 |
| Gamma | .862 | .695 | .588 | .581 | .546 | .21 |

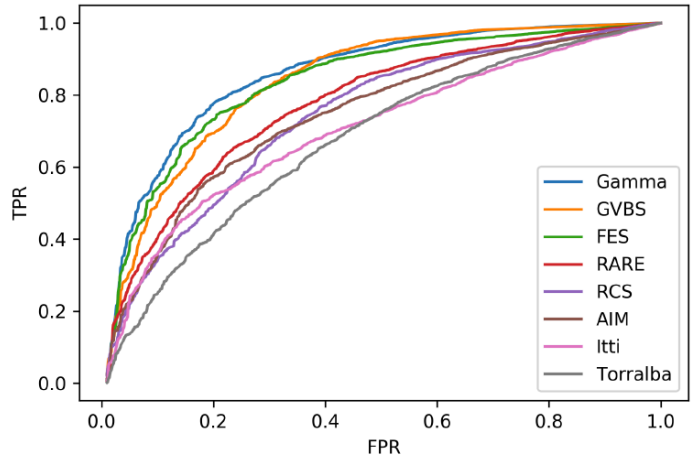

Figure 7. ROC curves on the Toronto Saliency Dataset. The images contain 3.8 times as many negative locations as positives.

Table 2 Attention Prediction Results on the CAT 2000 Database

| Method \ Metric | ROC (Judd) | ROC (Borji) | Similarity | Correlation | NSS | Time (s) |
|---|---|---|---|---|---|---|
| Itti | .700 | .570 | .377 | .206 | .258 | .25 |
| AIM | .772 | .628 | .437 | .335 | .497 | 1.04 |
| Torralba | .770 | .619 | .437 | .324 | .448 | 1.20 |
| GBVS | .844 | .642 | .498 | .486 | .510 | 1.05 |
| FES | .812 | .576 | .562 | .628 | .368 | .29 |
| RARE2012 | .822 | .643 | .466 | .408 | .511 | 1.37 |
| RCS | .763 | .593 | .431 | .292 | .352 | 14.91 |
| Gamma | .852 | .676 | .592 | .633 | .468 | .21 |

*Eye-Tracking Results*

In most eye tracking studies, users typically fixate on the center of the screen before the presentation of each visual stimulus. To best parallel human experiments, the present study begins the analysis of each image with the highest resolution (the artificial fovea) at the center of the screen. Table 3 shows the results for each saliency measure on the foveated Toronto database. Gamma saliency still performs the best across 5 of the 6 metrics, which shows that it could be used in a fixation system that approximates the HVS. Interestingly, when compared with Table 2, the foveation actually improves the results obtained by most saliency measures, possibly because the addition of a blur and center bias improves results, as shown in previous studies.

Table 3. Attention Prediction Results on the Foveated Toronto Database

| Method \ Metric | ROC (Judd) | ROC (Borji) | Similarity | Correlation | NSS |
|---|---|---|---|---|---|
| Itti | .737 | .597 | .403 | .314 | .369 |
| AIM | .794 | .657 | .433 | .458 | .561 |
| Torralba | .784 | .650 | .433 | .469 | .539 |
| GBVS | .839 | .664 | .502 | .603 | .594 |
| FES | .846 | .571 | .487 | .536 | .403 |
| RARE2012 | .841 | .656 | .525 | .632 | .591 |
| RCS | .819 | .629 | .517 | .595 | .517 |
| Gamma | .858 | .684 | .607 | .649 | .483 |

*Classification Results: Street View House Numbers*

The Street View House Numbers (SVHN) dataset offers a tough localization and classification challenge. It consists of over 73,000 training digits and over 23,000 testing digits in images from Google Street View. There are two main formats to the database - one cropped into 32x32 MNIST like digits with the additions of color, variable contrast, and some confusing data and the full images which contain extensive backgrounds and multiple digits in addition to the challenges in the cropped format.

In this dataset, the foveate vision implemented with Gamma saliency is used to localize and separate each number, turning the task into one resembling MNIST rather than training a single CNN to recognize both the number of digits and the classification of each. By using this divide-and-conquer approach the unsupervised feature extraction is able to focus on representing relevant parts of the image rather than trying to explain both the digit and the noise, leading to more useful features.

Figure 8 shows the foveate vision working on an example SVHN image. It first segments a salient area (Fig 8b-d), then breaks that area up into the digits that compose the two objects found in that location (Fig 8 e-j). By separating the digits in this manner, we are able to extract features that correspond to a single object at a time, rather than attempting to learn a network that explains an entire scene with multiple labeled objects.

Table 4 shows the classification accuracies, segmentation accuracies, and time per training epoch on the dataset with the enlarged bounding boxes created by the procedure outlined in [77]. Segmentation accuracies are calculated by dividing the intersection of the true and predicted bounding boxes by their union. The STN and CNN results are reported by the authors [24], [61] respectively, and do not include segmentation data or time information. We implemented an autoencoder [87] and a stacked autoencoder [88], which were not competitive. RWTA and TDN (the recurrent part of the RWTA is substituted by a time delay neural network [89], which greatly simplifies the training because it can be trained by backpropagation) both vastly outperform traditional unsupervised learning strategies that do not use attention. Once again there is only a small difference between the RWTA and TDN since the videos are artificially produced from still images. In addition, the focus of attention also improves the performance of a CNN close to the state-of-the-art results reported by the STN. This means that the focus of attention can be used with any neural network architecture for vision.

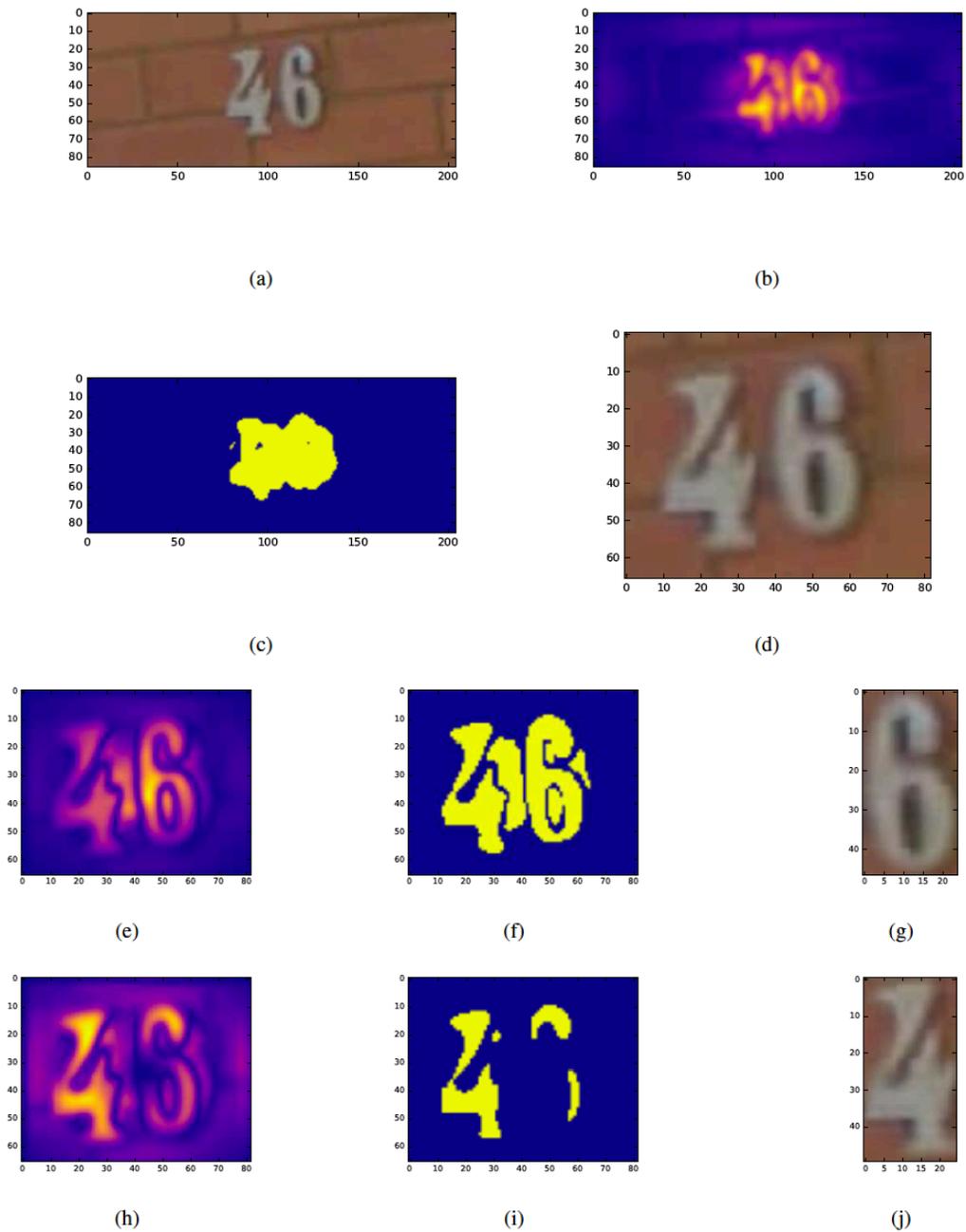

Figure 8 Initial SVHN image (A), saliency map produced by Gamma Saliency (B), thresholded saliency map (C), the cropped patch around the house numbers (D), the initial saliency map from the crop (E), the thresholded version of that map (F), the patch extracted around the object with the highest saliency (G), the saliency map with return inhibition around the most salient point (H), the thresholded map (I), and the cropped second object (J).

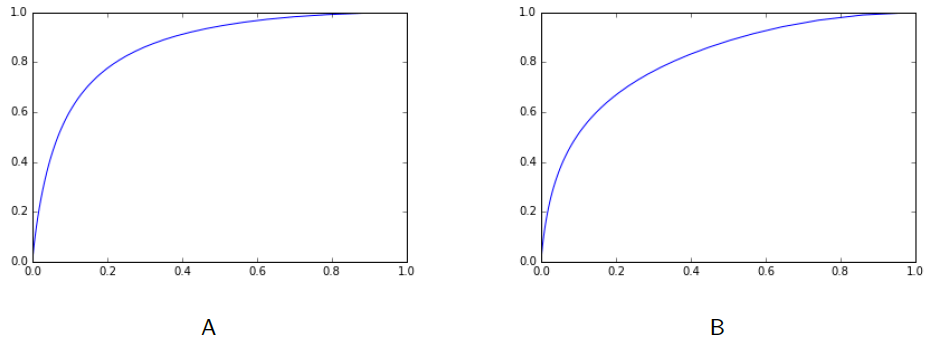

Figure 9. ROC curve for finding the bounding box containing all numbers from the SVHN data set using foveated vision (A) and without blurring and centering (B)

Figure 9 shows the ROC curves for the bottom-up attention system on the full SVHN dataset. These curves were created by setting the digit locations as fixations and computing the ROC using the Judd method [62].

TABLE III
SVHN RESULTS ON THE BOUNDED DATASET

| Method \ Metric | Unsupervised | | | | Supervised | | |
|---|---|---|---|---|---|---|---|
| | RWTA w/ Attention | TDN w/ Attention | Autoencoder | VAE | CNN Full | CNN FOA | STN Full |
| Classification | 92.51 | 92.28 | 15.58 | 17.46 | 94.47 | 96.06 | 96.30 |
| Segmentation | 83.67 | 83.67 | 76.92 | 76.92 | 76.92 | 83.67 | NA |
| Time (s) | 12638 | 2015 | 2372 | 2784 | 1426 | 1195 | NA |

TABLE IV
SVHN RESULTS ON THE FULL IMAGE DATASET

| Method \ Metric | Unsupervised | | | | Supervised | | |
|---|---|---|---|---|---|---|---|
| | RWTA w/ Attention | TDN w/ Attention | Autoencoder | VAE | CNN Full | CNN FOA | STN Full |
| Classification | 73.30 | 71.59 | 5.13 | 8.34 | 68.15 | 80.58 | 28.03 |
| Segmentation | 72.43 | 72.43 | NA | NA | NA | 72.43 | NA |
| Time (s) | 22658 | 2943 | 3689 | 4016 | 2397 | 1506 | 4549 |

Most results reported on the SVHN dataset use cropped digits, and even the few ones that try to classify the full address at once use an enlarged bounding box instead of the full image. In this next test, however, we use the full collected images with no additional preprocessing nor information about the image contents. This means that our bottom-up attention mechanism based on foveate vision implemented with the Gamma saliency must find the address location, segment, then identify each digit for success. This is a much harder problem than simply classifying boxed digits since it combines the problems of localization and classification in a paradigm that does not have fixed output size.

Table 4 shows the classification, segmentation, and timing results for the full dataset. In this case, adding an attention mechanism is imperative to success as the task involves classifying numbers in what are often extremely large background compared to the size of the numbers. It is gratifying to see that pure unsupervised techniques using space-time information are outperforming the standard deep architectures based on CNNs in curated datasets. This shows that the curated datasets used by the ML community are still very artificial and do not portray the reality of autonomous vision. It is also important to note that the foveate vision helps as a preprocessor for CNNs, because it avoids the cluttered background, even though there are still errors in the saliency algorithms.

*Top-down saliency in foveate vision: Occluded object recognition.*

In addition to testing on synthetic environments (cluttered MNIST [65]), we test the combined model saliency on a set of naturalistic images containing animals (Figure 10).

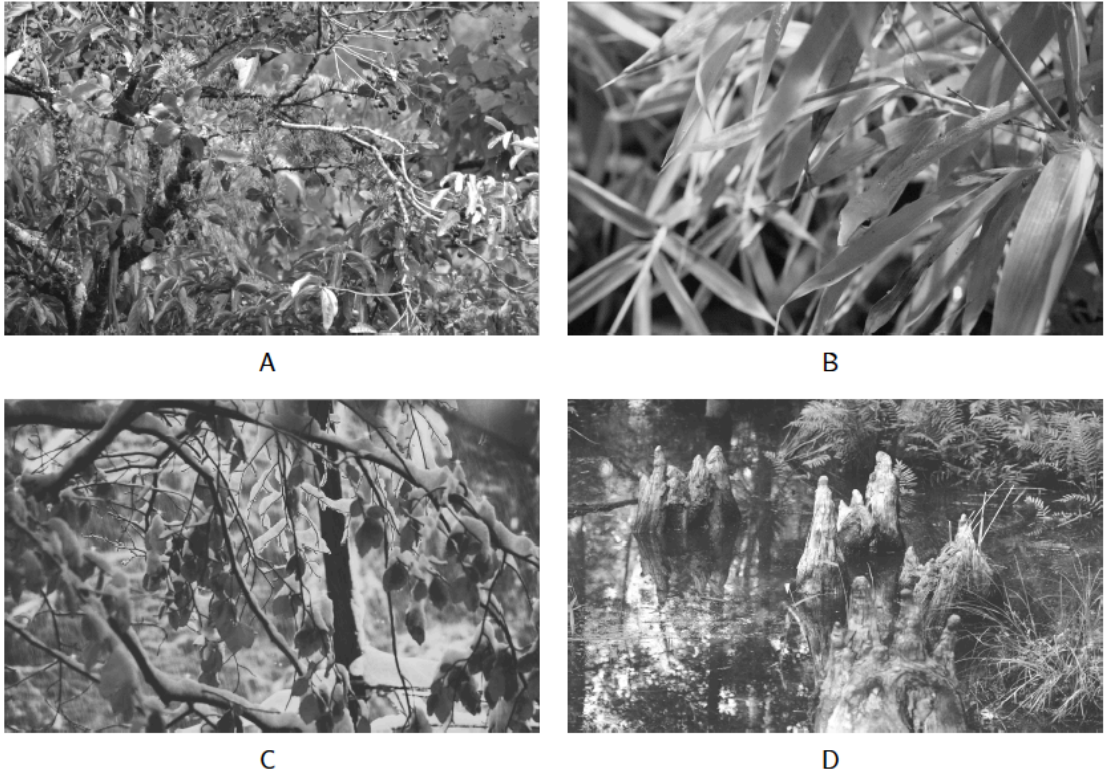

Figure 10. Example images from the naturalistic search dataset. An image containing a bird (A), a snake (B), an image with no bird (C), and an image with no snake (D).

<u>Experimental set up.</u> Thirty participants (U. of Florida students) were recruited. All participants reported normal or corrected-to-normal vision and no personal or family history of epilepsy or photic seizures. All participants provided informed consent, in accordance with the Declaration of Helsinki and the Institutional Review Board of the University of Florida.

Participants visually searched 80 nature scenes for either a hidden bird or snake. On each trial, a fixation cross was presented for a random duration between 1.5 and 3 seconds (square window), followed by a target cue depicting either a bird of snake, indicating which hidden animal participants should search for in this trial (Figure 11). Target cues had a random duration between 1.5 and 3 seconds (square window) and were followed by a high-resolution image of a nature scene. Participants were instructed that they had 15 seconds to respond to each picture, by pressing one button to indicate they found the hidden target, or another button to indicate no target was hidden in the picture. If the participant made no response in the allocated 15 seconds, the stimulus automatically offset, and the next trial began.

Both the fixation cross and the target cue were white against a black background, had the same mean luminance as the images, and occupied a visual angle of 5.71°. Each scene was controlled for luminance and color. Luminance was controlled for by adjusting the mean brightness of each picture to have a mean luminance of 31 cd/m$^2$. Color was controlled for by converting each image to greyscale. To elicit a neural signal representative of visual attention, the all pixels in the scene flickered at 30 Hz. This frequency was chosen as its high enough so as not to be perceptually distracting (it looks like an old movie), but low enough to still produce an acceptable signal-to-noise ratio.

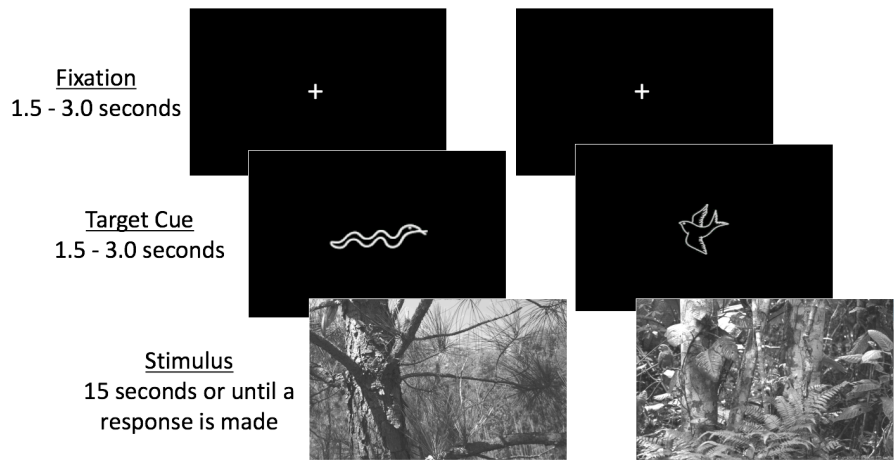

Figure 11. Trial example. Each trial consisted of a fixation cross, followed by a target cue, and ending with a nature scene. Target cues indicated whether a participant should search for either a bird or a snake in the subsequent picture. Of the eighty natural scenes, half were preceded by a bird cue, and the other half were preceded by a snake cue to indicate which target the participant should search for. Targets were present in half of the trials and always matched the cue proceeding it.

Participants' fixations and eye-movements were recorded during the experiment using an EyeLink 1000 plus eye tracker using a 16mm lens. Pupil size and gaze location were both sampled at 500 Hz. Participants were comfortably seated 1.3 meters from the monitor such that the illuminator was parallel with the lower edge of the monitor. The illumination level of the infrared signal was set to 100% and based were adjusted on an individual basis for the pupil and corneal reflection. Participants were asked to minimize head-movements and jaw-movements for the duration of the experiment and used a chin rest to help them keep still. Gaze location was calibrated for each participant before the experiment began. During the calibration, the lens and pupil thresholds were adjusted while participants viewed a white circle (5° visual angle) at 9 different locations on a black screen. This task was repeated until the pupil and gaze location could be identified at each of the 9 points.

The mean time taken by 30 human subjects to find a target was 6.5 sec. (sd = 3.8 sec), and the number of correct responses was 70.5% (sd = 9.8%). In target trials, the mean number of saccades to find the target (bird, snake) was 16.9, (range = 3 - 30, median = 15.5, sd = 9.1).

We provided the same information to our vision system through the top down-input, by weighting online the bottom up mask for the saliency with the information obtained from the training set for the animal that was going to appear in the image. This mimics the top down attention that is known to exist in humans. Since we are dealing with such a small number of images it would be impractical to train an entire neural network to differentiate birds from snakes from background. Therefore, we implemented transfer learning [90] with the feature extractor from a network pre-trained on a much larger dataset. For this task, we use the VGG16 network [91] which is a 16 layers network with 13 convolutional layers and three fully-connected layers trained on Imagenet [1]. The output of the final convolutional layer (before the max-pooling) will be used to generate the feature maps used as inputs to the top-down saliency. This produces 512 feature maps.

To test the added value of using top-down and bottom up saliency measures over the sole use of the bottom-up saliency measure, first we used each algorithm to find the animal in the 40 images that contain a target. The target is considered found if the image patch produced by the saliency map has an intersection over union of greater than .5 with the true bounding box. The patches are obtained using the method described in section IIID without the extra augmentation used to create videos for the RWTA. The weights for the maps in the top-down saliency were created by randomly selecting an image with a bird and another with a snake. Both the top-down and bottom-up saliencies are computed using $k = (1, 60, 1, 38, 1, 19)$ and $\mu = (.05, .5, .1, .5, .5, .5)$, with $n_1 = n_2 = 400$. Across the 38 remaining images that contain targets, the bottom-up saliency finds the target in an average of 39.8 saccades, while the top-down saliency finds the target in an average of 5.9 saccades. The top-down saliency both finds the targets faster and produces much more realistic maps, and in fact outperforms the human subjects in the number of saccades necessary to find the animal in target trials. Figure 12 shows an example image containing an occluded bird along with the final bottom-up and top-down saliency maps. The bottom-up saliency map highlights the edges of the central objects fairly equally, while the top down saliency focuses on the bird even though it is partially hidden by leaves in the foreground. It only takes 4 fixations to find the bird for the top-down saliency, while the bottom-up saliency takes 16 fixations for this image.

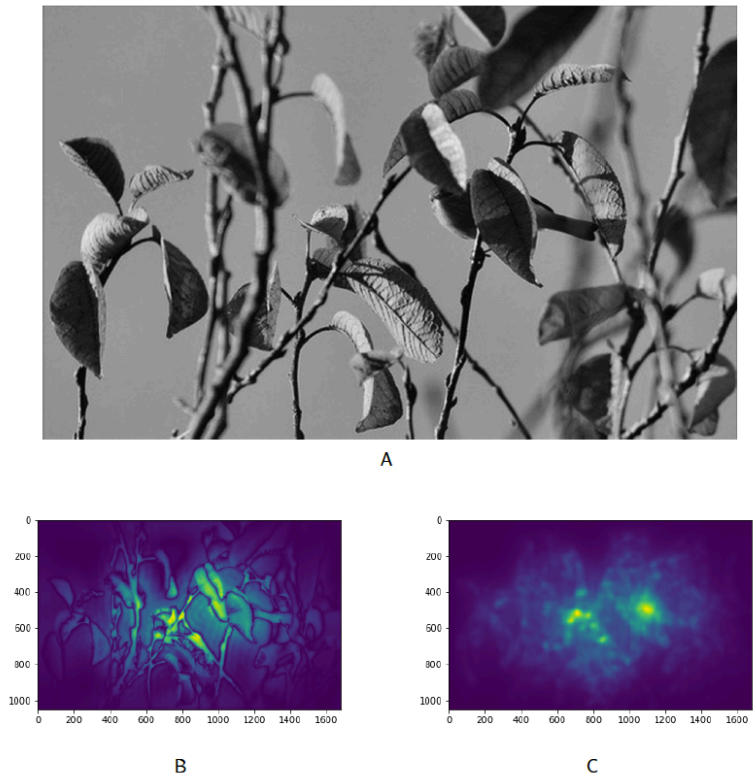

Figure 12. Top-down vs. bottom-up saliency on the naturalistic dataset. An image containing a bird (A), the bottom-up saliency map (B), and the top-down saliency map (C).

In addition to how quickly the targets are found, we can analyze the top-down search by how well it matches a human searching the same scene. From this eye-tracking data, we can create fixation maps that can be compared to saliency maps. Fig 13 shows an example image with the eye-tracking data of a single person overlaid. The person scans most of the image before finding the target rather than localizing it instantly, showing that this is a hard task even for humans. For the same image the foveal vision system will compute a sequence of saliency maps until it finds the animal, so we also know the sequence and number of "saccades" the computer did to find the animal. Fig 13 shows one example with the bottom-up salience and the combined bottom-up and top-down saliency map.

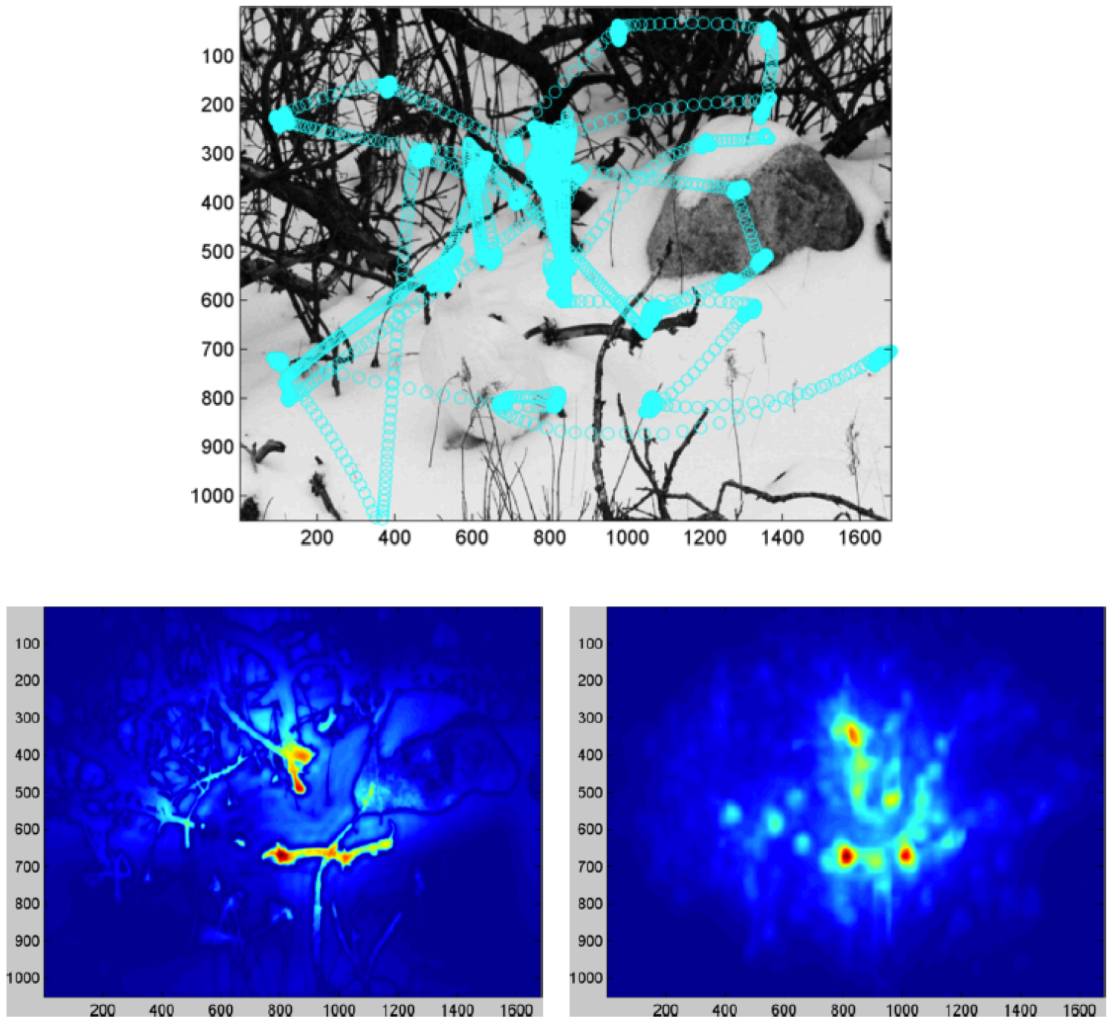

Figure 13. Search image with overlaid eye-tracking along with the associated bottom-up (left) and top-down + bottom up (right) maps.

Since the participants in the study were told to find the animals, we expect the top-down saliency to outperform the bottom-up saliency because we are specifically biasing it towards finding the targets. Table 5 shows the results of the two saliency measures with the commonly used metrics. It does not show how successful either bottom-up or top-down saliency are at finding the targets, but how well the fixation locations match the locations searched by humans.

Table 5. Comparison of Human Search with Top-Down and Bottom-Up Saliency

| Method \ Metric | ROC (Judd) | ROC (Borji) | Similarity | Correlation | NSS |
|---|---|---|---|---|---|
| Bottom-Up | .586 | .427 | .362 | .267 | .398 |
| Top-Down | .792 | .615 | .562 | .331 | .461 |

Figure 14 shows the ROC curve for each saliency. The top-down saliency does outperform the bottom-up version greatly in all of the metrics, showing that the top-down saliency does approximate human search better than the purely bottom-up approach. Bottom-up saliency measures have been shown to correlate with free-exploration in images [26], but fail when the fixation maps are produced by humans that are searching for specific objects. The top-down saliency mitigates this problem by using features to search for these targets. Doing this not only finds the targets in less fixations, but searches the images in a way that more closely matches human exploration than the bottom-up method.

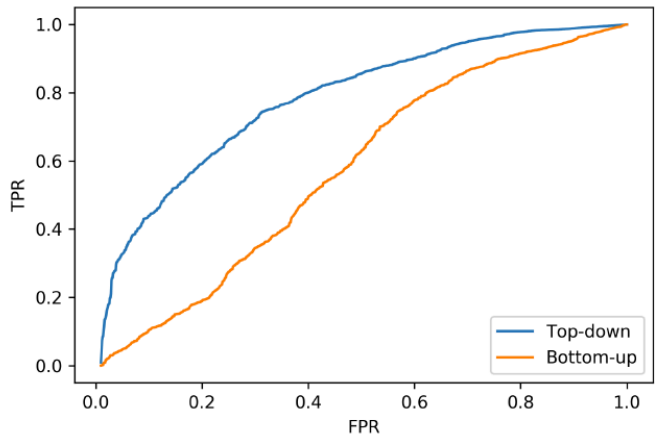

Figure 14. ROC curves for top-down and bottom-up saliency for a search task. The images contain 5.4 times more negative locations than positives.

V. CONCLUSION

In this paper we propose an architecture that mimics the function of two fundamental mechanisms of the human vision system: a saliency-based method for the spatial attention (approximating functions of the HVS dorsal pathway) and an object recognition network (approximating the function of the HVS ventral pathway). By separating these pathways, we can achieve greater computational efficiency by quickly selecting subregions of the image for full processing, as well as improve the feature extraction by eliminating non-discriminatory data.

In addition to removing irrelevant information, the attention pathway is a way to also create videos from still images, which adds data augmentation to the new architecture and in spite of being unsupervised, gets performance close to fully supervised techniques. Here we use a RTWA network that has both spatial and temporal pathways to learn more discriminant data structure. This combination extracts robust features that cluster the relevant information in objects without the need of labels. One major drawback to spatial time features is the computation time, especially the temporal component that uses an RNN. To mitigate, we simply combined the features from different frames using learned weights (TDN), which produced similar results at much faster speeds. We believe this was possible since the underlying data in the frames was the same, i.e. the videos were created simply with rotation. In a true video, the RNN would be necessary to learn the temporal relationship between the frames.

The inclusion of the top-down saliency is a major advantage in conditions of high background clutter, and in scene understanding with many objects. Fovea vision (Fig 3) must go over all of the salient points which can be very computation intensive and the visitation order is dictated by pixel information only. Top-down attention is able to change this ordering, when the information is made available to the system. Basically, the vision system must use fovea vision to individualize each object, and store the extracted object features in an external memory (content addressable memory – CAM) as shown in Fig 2. This problem requires the use of extra information to disambiguate the images, which the brain does naturally through attention mechanisms. We tested this method with question and answer problems [ref], but here we compared the system performance with human observers in naturally cluttered images. Since we had to use a traditional CNN trained on a large dataset, we had to invent a way to extract objects specific units in the top layers of CNNs. Our proposal of using saliency applied to the top layers of a trained deep learning architecture is simple but very effective, and it shows that the local activation of units in the top layer of CNNs indeed carry global information about the image type in a distributed manner. The issue is to find where the local information resides in the layers. This is where the concept of Gamma saliency shines, because it can select both the area and its size to maximize the "contrast". We only use a linear model to make the decision, and nonlinear functions would likely improve the results. Very few computer vision systems today could achieve the performance of our method bottom-up and top-down vision architecture. This is an exciting avenue of future research.

Future work includes extending and testing the proposed method on video. The dual spatial-temporal architecture of the RWTA make it uniquely suited to extracting features from videos, so pairing it with an appropriate attention mechanism that takes time into account should learn robust features. In addition, we could research improving the attention mechanism to make it focus not only on areas of the image that are locally different, but ones that offer the greatest scene understanding when combined with the information already extracted from the image.

Acknowledgement: This work was partially funded by ONR-N00014-14-1-0542, N00014-18-1-2306 and DARPA FA9453-18-1-0039.

## VI. APPENDIX A

### A. Cluttered MNIST

All networks were created using Keras [92] using the Theano backend. We adopt the notation that conv[N,w,s] denotes a convolutional layer with N filters of size wxw, and stride s; fc[N] is a fully connected layer with N units; and max[s] is a sxs max-pooling layer with stride s.

The CNN model is conv[64,9,9]-max[2]- conv[32,7,7]-max[2]-fc[256]-fc[10] with rectified linear units following each weight layer and a softmax layer at the end for classification. For the CNN with STN, the STN network is max[2]- conv[20,5,5]-max[2]-conv[20,5,5]-fc[50]-fc[6].

The spatial encoder in the RWTA model is conv[64,3,3]-conv[64,3,3], while the convolutional time encoder is conv[64,3,3] with a time sequence of 5 frames (Figure A). A linear SVM is learned on the latent states of the RWTA to produce the classification scores.

Since the images and digits in this dataset are uniformly sized, a single scale attention model was used. The center kernel has an order of k = 1 and a shape parameter of μ = 0.2. The neighborhood kernel has an order of k = 9 and μ = 0.5. A single frame was extracted from each image since each image contained only a single digit and contained no location information. Each network was trained for 500 epochs on a Tesla K80 GPU.

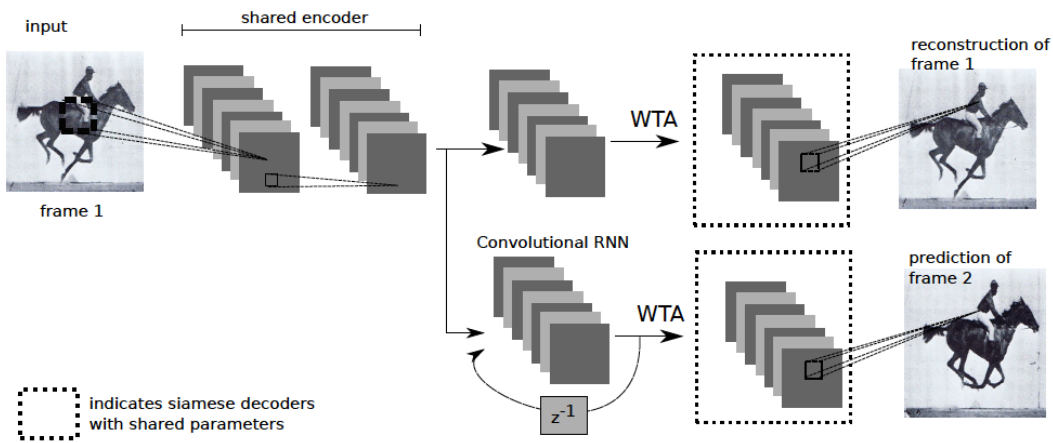

Figure A. Block diagram of the RTWA.

### B. SVHN

The CNN model is: conv[48,5,1]-max[2]- conv[64,5,1]-conv[128,5,1]-max[2]-conv[160,5,1]-conv[192,5,1]-max[2]-conv[192,5,1]-conv[192,5,1]- max[2]-conv[192,5,1]-fc[3072]-fc[3072]-fc[3072], with rectified linear units following each weight layer, followed by five parallel fc[11] and softmax layers for classification. There are 11 outputs in the final layer to account for the digits 0-9 and an extra class for noise classification. The ST-CNN has a single spatial transformer before the first convolutional layer of the CNN model the STNs localization network architecture is: conv[32,5,1]-max[2]-conv[32,5,1]-fc[32]-fc[32].

The spatial encoder in the RWTA model is conv[64,3,3]-conv[64,3,3]-conv[128,3,3], while the convolutional time encoder is conv[64,3,3] with a time sequence of 5 frames. A linear SVM is learned on the latent states of the RWTA to produce the classification scores.

Since the images and digits in this dataset have different sizes, a multi scale attention model was used. The center kernels has an order of k = 1; k = 1; k = 1 and a shape parameter of $\mu$ = 0.1; $\mu$ = 0.3; and $\mu$ = 0.8. The neighborhood kernel has an order of k = 13; k = 9; k = 5 and $\mu$ = 0.3; $\mu$ = 0.5; $\mu$ = 0.7. These parameters were used to create the initial saliency maps and find the main fixation points. For the local saliency, the largest scale was removed to focus on finer details, leaving a two-scale kernel. Each network was trained for 10000 epochs on a Tesla K80 GPU.